\documentclass{article}

\usepackage[final]{nips_2017}

\usepackage[utf8]{inputenc} 
\usepackage[T1]{fontenc}    
\usepackage{url}            
\usepackage{booktabs}       
\usepackage{amsfonts}       
\usepackage{nicefrac}       
\usepackage{microtype}      
\usepackage{natbib}
\setcitestyle{authoryear,open={(},close={)}}
\usepackage[pdftex]{graphicx}

\usepackage{graphicx,picture}
\usepackage{mathtools}
\usepackage{amssymb,amsmath,amsthm}

\usepackage{subfigure}
\usepackage{sidecap}

\usepackage{tikz}
\def\checkmark{\tikz\fill[scale=0.4](0,.35) -- (.25,0) -- (1,.7) -- (.25,.15) -- cycle;} 

\usepackage{placeins}

\def\argmax{\mathop{\arg\max}}

\newtheorem{definition}{Definition}

\def\E{{\mathbb E}} 
\def\R{{\cal R}}

\def\R{{\cal R}}

\def\bs{{\mathbf{s}}}
\def\ba{{\mathbf{a}}}

\definecolor{darkgreen}{RGB}{0,125,0}

\title{Value-Decomposition Networks For Cooperative Multi-Agent Learning}

\author{
  Peter Sunehag\\DeepMind\\sunehag@google.com \And Guy Lever\\DeepMind\\guylever@google.com \And Audrunas Gruslys\\DeepMind\\audrunas@google.com \And Wojciech Marian Czarnecki\\DeepMind\\lejlot@google.com \And Vinicius Zambaldi\\DeepMind\\vzambaldi@google.com
  \And Max Jaderberg\\DeepMind\\jaderberg@google.com \And Marc Lanctot\\DeepMind\\lanctot@google.com \And Nicolas Sonnerat\\DeepMind\\sonnerat@google.com \And Joel Z. Leibo\\DeepMind\\jzl@google.com \And Karl Tuyls\\DeepMind \& University of Liverpool\\karltuyls@google.com \And Thore Graepel\\DeepMind\\thore@google.com 
}

\begin{document}

\maketitle

\begin{abstract}
We study the problem of cooperative multi-agent reinforcement learning with a single joint reward signal. This class of learning problems is difficult because of the often large combined action and observation spaces. In the fully centralized and decentralized approaches, we find the problem of spurious rewards and a phenomenon we call the ``lazy agent'' problem, which arises due to partial observability.  We  address these problems by training individual agents with a novel value decomposition network architecture, which learns to decompose the team value function into  agent-wise value functions. We perform an experimental evaluation across a range of partially-observable multi-agent domains and show that learning such value-decompositions leads to superior results, in particular when combined with weight sharing, role information and information channels. 
\end{abstract}

\section{Introduction}

We consider the cooperative multi-agent reinforcement learning (MARL) problem \citep{panait05,Busoniu08MARL,TuylsW12}, in which a system of several learning agents must jointly optimize a single reward signal -- the \emph{team reward} -- accumulated over time. Each agent has access to its own (``local'') observations and is responsible for choosing actions from its own action set. Coordinated MARL problems emerge in applications such as coordinating self-driving vehicles and/or traffic signals in a transportation system, or optimizing the productivity of a factory comprised of many interacting components. More generally, with AI agents becoming more pervasive, they will have to learn to coordinate to achieve common goals.

Although in practice some applications may require local autonomy, in principle the cooperative MARL problem could be treated using a \emph{centralized} approach, reducing the problem to single-agent reinforcement learning (RL) over the concatenated observations and combinatorial action space. We show that the centralized approach consistently fails on relatively simple cooperative MARL problems in practice. We  present a simple experiment in which the centralised approach fails by learning inefficient policies with only one agent active and the other being ``lazy''. This happens when one agent learns a useful policy, but a second agent is discouraged from learning because its exploration would hinder the first agent and lead to worse team reward.\footnote{For example, imagine training a 2-player soccer team using RL with the number of goals serving as the team reward signal. Suppose one player has become a better scorer than the other. When the worse player takes a shot the outcome is on average much worse, and the weaker player learns to avoid taking shots \citep{HausknechtThesis}.} 

An alternative approach is to train \emph{independent learners} to optimize for the team reward. In general each agent is then faced with a non-stationary learning problem because the dynamics of its environment effectively changes as teammates change their behaviours through learning \citep{Laurent11}. Furthermore, since from a single agent's perspective the environment is only partially observed, agents may receive spurious reward signals that originate from their teammates' (unobserved) behaviour. Because of this inability to explain its own observed rewards naive independent RL is often unsuccessful: for example \citet{ClausBoutillierDynamics} show that independent $Q$-learners cannot distinguish teammates' exploration from stochasticity in the environment, and fail to solve even an apparently trivial, 2-agent, stateless, $3\times 3$-action problem and the general Dec-POMDP problem is known to be intractable \citep{BernsteinDecPomdp,OliehoekAmato16book}. Though we here focus on 2 player coordination, we note that the problems with individual learners and centralized approaches just gets worse with more agents since then, most rewards do not relate to the individual agent and the action space grows exponentially for the fully centralized approach.

One approach to improving the performance of independent learners is to design individual reward functions, more directly related to individual agent observations. However, even in the single-agent case, reward shaping is difficult and only a small class of shaped reward functions are guaranteed to preserve optimality w.r.t. the true objective \citep{NgShaping,tumer-devlin_aamas14,EckSDK16}. In this paper we aim for more general autonomous solutions, in which the decomposition of the team value function is learned.


We introduce a novel {\bf learned additive value-decomposition} approach over individual agents. 
Implicitly, the value decomposition network aims to learn an optimal linear value decomposition from the team reward signal, by back-propagating the total $Q$ gradient through deep neural networks representing the individual component value functions. 
This additive value decomposition is specifically motivated by avoiding the spurious reward signals that emerge in purely independent learners.
The implicit value function learned by each agent depends only on local observations, and so is more easily learned. 
Our solution also ameliorates the coordination problem of independent learning highlighted in \citet{ClausBoutillierDynamics} because it effectively learns in a centralised fashion at training time, while agents can be deployed individually. 

Further, in the context of the introduced agent, we evaluate weight sharing, role information and information channels as additional enhancements that have recently been reported to improve sample complexity and memory requirements \citep{HausknechtThesis,FoersterCommunicate, SF16}. However, our main comparison is between three kinds of architecture; Value-Decomposition across individual agents, Independent Learners and Centralized approaches. We investigate and benchmark combinations of these techniques applied to a range of new interesting two-player coordination domains. We find that Value-Decomposition is a much better performing approach than centralization or fully independent learners, and that when combined with the additional techniques, results in an agent that consistently outperforms centralized and independent learners by a big margin.

\subsection{Other Related Work}

\citet{SchneiderDistributed} consider the optimization of the sum of individual reward functions, by optimizing local compositions of individual value functions learnt from them. \citet{RussellZimdarsQDecomposition} sums the $Q$-functions of independent learning agents with individual rewards, before making the global action selection greedily to optimize for total reward. Our approach works with only a team reward, and \emph{learns} the value-decomposition autonomously from experience, and it similarly differs from the approach with coordination graphs \citep{Guestrin2002} and the max-plus algorithm \citep{KuyerWBV08,Pol16}.


Other work addressing team rewards in cooperative  settings is based on \textit{difference rewards} \citep{tumer-wolpert_cdcs04}, measuring the impact of an agent's action on the full system reward. This reward has nice properties (e.g. high learnability), but can be impractical as it requires knowledge about the system state \citep{tumer-colby_aamas16,tumer-agogino_jaamas08,tumer-proper_aamas12}. Other approaches can be found in \citet{tumer-devlin_aamas14,tumer-holmesparker_ker14,BabesCL08}.



\section{Background}

\subsection{Reinforcement Learning}

We recall some key concepts of the RL setting \citep{SB98}, an agent-environment framework \citep{RN10} in which an agent sequentially interacts with the environment over a sequence of timesteps, $t=1,2,3,\hdots $, by executing actions and receiving observations and rewards, and aims to maximize cumulative reward. This is typically modelled as a Markov decision process (MDP) \citep[e.g.][]{puterman} defined by a tuple $ \langle \mathcal{S,A},\mathcal{T}_1,\mathcal{T},R \rangle$ comprising the state space $\mathcal{S}$, action space $\mathcal{A}$, a (possibly stochastic) reward function $R : \mathcal{S \times A\times S }\to \mathbb{R}$ start state distribution $\mathcal{T}_1 \in \mathcal{P(S)}$ and transition function $\mathcal{T}:\mathcal{S\times A} \rightarrow \mathcal{P(S)}$, where $\mathcal{P(X)}$ denotes the set of probability distributions over the set $\mathcal{X}$. We use $\bar{R}$ to denote the expected value of $R$. The agent's interactions give rise to a trajectory $(S_1,A_1,R_1,S_2,...)$ where $S_{1}\sim \mathcal{T}_1$, $S_{t+1}\sim \mathcal{T}(\cdot|S_t,A_t)$ and $R_t = R(S_t,A_t,S_{t+1})$, and we denote random variables in upper-case, and their realizations in lower-case. At time $t$ the agent observes $ o_t\in\mathcal{O} $ which is typically some function of the state $s_t$, and when the state is not fully observed the system is called a partially observed Markov decision process (POMDP).

The agent's goal is to maximize expected cumulative discounted reward with a discount factor $\gamma$, $\R_t := \sum_{t=1}^{\infty}{\gamma^{t-1} R_{t}}$. The agent chooses actions according to a \emph{policy}: a (stationary)  policy is a function $\pi:\mathcal{S}\to\mathcal{P(A)}$ from states to probability distributions over $\mathcal{A}$. An optimal policy is one which maximizes expected cumulative reward. In fully observed environments, stationary optimal policies exist. In partially observed environments, the policy usually incorporates past agent observations from the \emph{history} $h_t=a_1o_1r_1,...,a_{t-1}o_{t-1}r_{t-1}$ (replacing $s_t$). A practical approach utilized here, is to parameterize policies using recurrent neural networks.

$V^\pi(s) := \mathbb{E} [\sum_{t=1}^\infty \gamma^{t-1} R(S_t, A_t,S_{t+1})|S_1 = s;A_t\sim \pi(\cdot|S_t)]$ is the value function and the action-value function is $Q^\pi(s,a) :=  \mathbb{E}_{S'\sim \mathcal{T}(\cdot|s,a)} [R(S, a,S') + \gamma V(S')]$ (generally, we denote the successor state of $s$ by $s'$). The optimal value function is defined by $V^*(s) = \sup _\pi  V^\pi(s)$ and similarly  $Q^*(s,a) = \sup_\pi Q^\pi(s,a)$. For a given action-value function $Q : \mathcal{S} \times \mathcal{A}\to \mathbb{R}$ we define the (deterministic) greedy policy w.r.t. $Q$ by $\pi(s) := \argmax_ {a\in\mathcal{A}} Q(s, a)$ (ties broken arbitrarily). The greedy policy w.r.t. $Q^*$ is optimal \citep[e.g.][]{SzepesvariAlgorithms}.

\subsection{Deep $Q$-Learning}

One method for obtaining $Q^*$ is $Q$-learning which is based on the update $Q_{i+1}(s_t,a_t)=(1-\eta_t) Q_i(s_t,a_t)+\eta_t(r_t+\gamma \max_a Q_i(s_{t+1},a))$, where $\eta_t\in (0,1)$ is the learning rate. We employ the $\varepsilon$-greedy approach to action selection based on a value function, which means that with $1-\varepsilon$ probability we pick $\argmax_a Q_i(s,a)$ and with probability $\varepsilon$ a random action. Our study focuses on deep architectures for the value function similar to those used by \citet{dqn15}, and our approach incorporates the key techniques of target networks and experience replay employed there, making the update into a stochastic gradient step. Since we consider partially observed environments our $Q$-functions are defined over agent observation histories, $Q(h_t,a_t)$,  and we incorporate a recurrent network similarly to \citet{HausknechtStoneRecurrent}. To speed up learning we add the dueling architecture of \citet{WangDuelling} that represent $Q$ using a value and an advantage function, including multi-step updates with a forward view eligibility trace \citep[e.g.][]{HarbPrecupRecurrance} over a certain number of steps. When training agents the recurrent network is updated with truncated back-propagation through time (BPTT) for this amount of steps. Although we concentrate on DQN-based agent architectures, our techniques are also applicable to policy gradient methods such as A3C \citep{MnihAsynchronous}.

\subsection{Multi-Agent Reinforcement Learning}

We consider problems where observations and actions are distributed across $d$ agents, and are represented as $d$-dimensional tuples of primitive observations in $\mathcal{O}$ and actions in $\mathcal{A}$. As is standard in MARL, the underlying environment is modeled as a Markov game where actions are chosen and executed simultaneously, and new observations are perceived simultaneously as a result of a transition to a new state ~\citep{Littman94,Littman01,HuW03,Busoniu08MARL}.

Although agents have individual observations and are responsible for individual actions, each agent only receives the joint reward, and we seek to optimize $\R_t$ as defined above. This is consistent with the Dec-POMDP framework \citep{OliehoekSV08, OliehoekAmato16book}. 

If we denote $\bar h := (h^1,h^2,...,h^d)$ a tuple of agent histories, a joint policy is in general a map $\pi: \mathcal{H}^d \to \mathcal{P}(\mathcal{A}^d)$; we in particular consider policies where for any history $\bar{h}$, the distribution $\pi(\bar{h})$ has independent components in $\mathcal{P}(\mathcal{A})$. Hence, we write  $\pi:\mathcal{H}^d\to \mathcal{P}(\mathcal{A})^d$. The exception is when we use the most naive centralized agent with a combinatorial action space, aka joint action learners. 


\section{A Deep-RL Architecture for Coop-MARL}

\begin{figure}
\begin{minipage}{.5\textwidth}
\centering
\includegraphics[width=3.5cm]{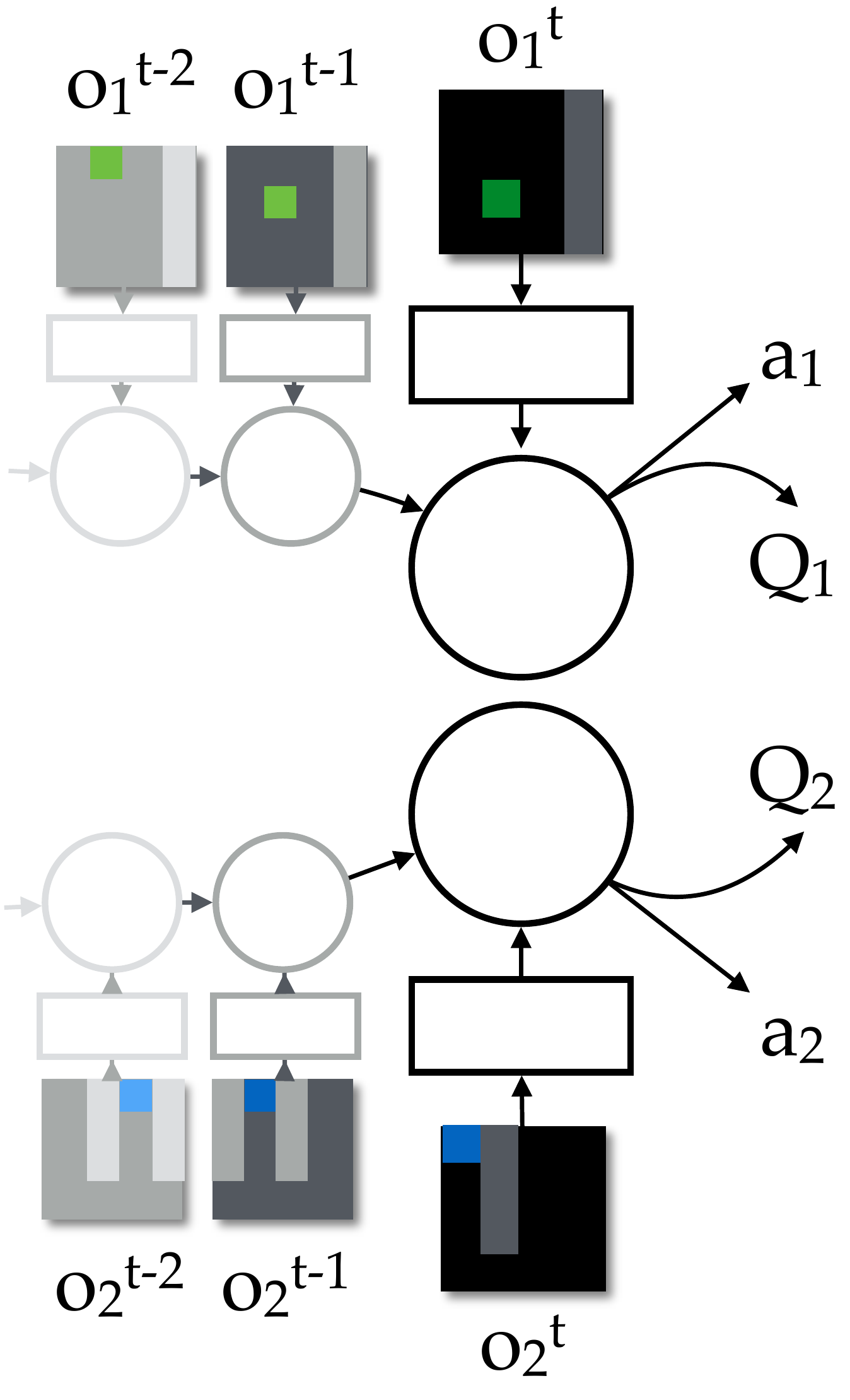}
\caption{Independent agents architecture showing how local observations enter the networks of two agents over time (three steps shown), pass through the low-level linear layer to the recurrent layer, and then a dueling layer produces individual $Q$-values.}
\label{Independent}
\end{minipage}
\begin{minipage}{.5\textwidth}
 \centering
\includegraphics[width=3.5cm]{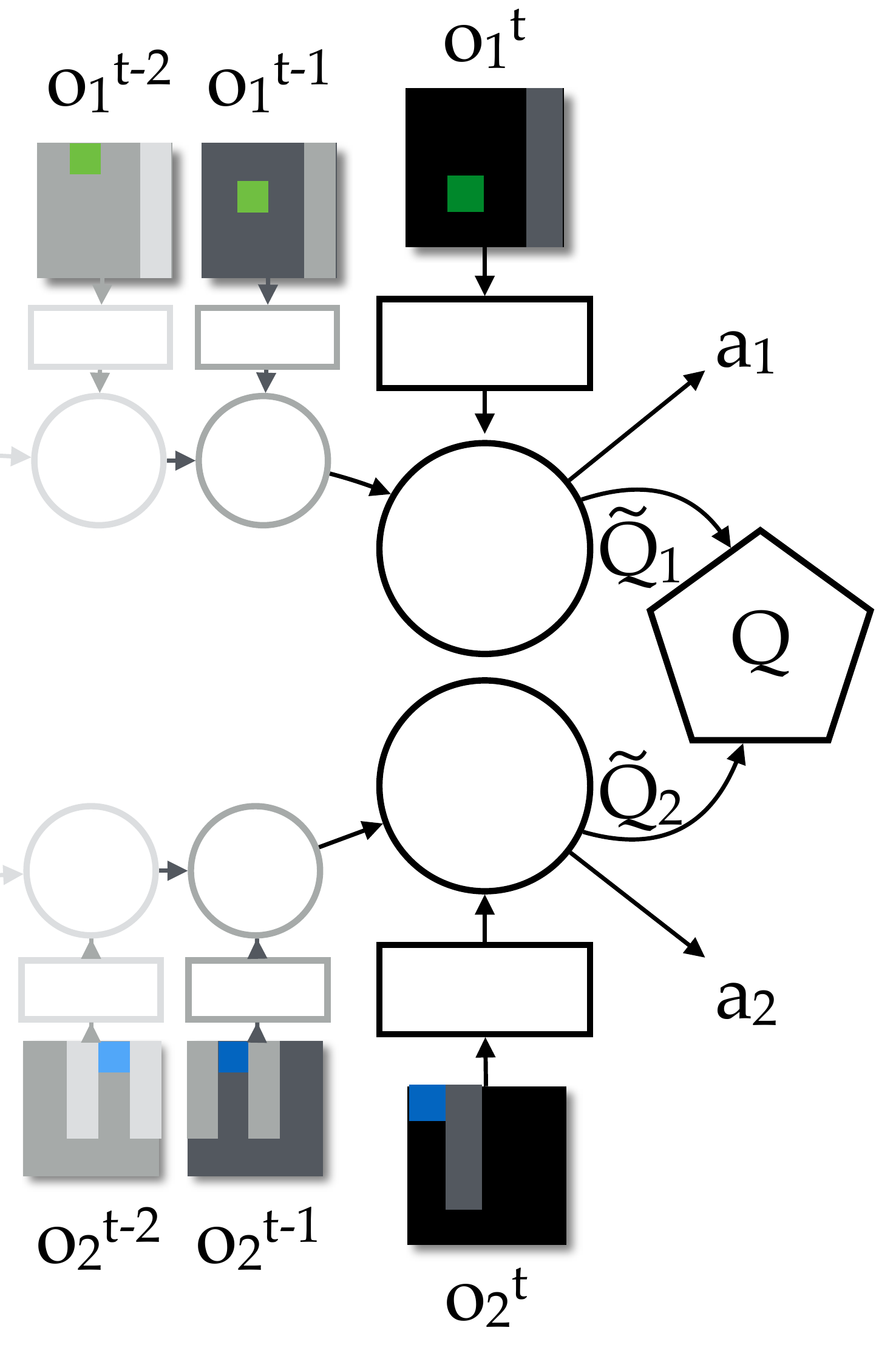}
\caption{Value-decomposition individual architecture showing how local observations enter the networks of two agents over time (three steps shown), pass through the low-level linear layer to the recurrent layer, and then a dueling layer produces individual "values" that are summed to a joint $Q$-function for training, while actions are produced independently from the individual outputs.}
\label{Value-Decomp}
\end{minipage}
\end{figure}

Building on purely independent DQN-style agents (see Figure~\ref{Independent}), we add enhancements to overcome the identified issues with the MARL problem. Our main contribution of value-decomposition is illustrated by the network in Figure~\ref{Value-Decomp}.

The main assumption we make and exploit is that the joint action-value function for the system can be additively decomposed into value functions across agents,
$$
Q((h^1,h^2,...,h^d),(a^1,a^2,...,a^d))\approx \sum_{i=1}^d \tilde Q_i(h^i,a^i)\label{Qsum}
$$
where the $\tilde Q_i$ depends only on each agent's local observations. We learn $\tilde Q_i$ by backpropagating gradients from the $Q$-learning rule using the joint reward through the summation, i.e. $\tilde Q_i$ is learned implicitly rather than from any reward specific to agent $i$, and we do not impose constraints that the $\tilde Q_i$ are action-value functions for any specific reward. The value decomposition layer can be seen in the top-layer of Figure~\ref{Value-Decomp}. One property of this approach is that, although learning requires some centralization, the learned agents can be deployed independently, since each agent acting greedily with respect to its local value $\tilde Q_i$ is equivalent to a central arbiter choosing joint actions by maximizing the sum $\sum_{i=1}^d \tilde Q_i$.


For illustration of the idea consider the case with 2 agents (for simplicity of exposition) and where rewards decompose additively across agent observations\footnote{Or, more generally, across agent histories.}, $r(\bs,\ba)=r_1(o^1,a^1)+r_2(o^2,a^2)$, where $(o^1, a^1)$ and $(o^2, a^2)$ are (observations, actions) of agents 1 and 2 respectively. This could be the case in team games for instance, when agents observe their own goals, but not necessarily those of teammates. In this case we have that
$$
Q^\pi(\bs,\ba) = \E [ \sum_{t=1}^\infty \gamma^{t-1} r(\bs_t,\ba_t) | \bs_1 = \bs, \ba_1 = \ba; \pi   ] \nonumber $$ $$
=\E [ \sum_{t=1}^\infty \gamma^{t-1} r_1(o^1_t,a^1_t) | \bs_1 = \bs, \ba_1 = \ba; \pi   ] + \E [ \sum_{t=1}^\infty \gamma^{t-1} r_2(o^2_t,a^2_t) | \bs_1 = \bs, \ba_1 = \ba; \pi   ] \nonumber $$ $$
 =: \bar{Q}^\pi_1(\bs,\ba) + \bar{Q}^\pi_2(\bs,\ba) \nonumber
$$
where $\bar{Q}^\pi_i(\bs,\ba):= \E [ \sum_{t=1}^\infty \gamma^{t-1} r_1(o^i_t,a^i_t) | \bs_1 = \bs, \ba_1 = \ba; \pi   ], i=1,2$. 
The action-value function $\bar{Q}^\pi_1(\bs,\ba)$ -- agent 1's expected future return -- could be expected to depend more strongly on observations and actions $(o^1, a^1)$ due to agent 1 than those due to agent 2. If $(o^1, a^1)$ is not sufficient to fully model $\bar{Q}^\pi_1(\bs,\ba)$ then agent 1 may store additional information from historical observations in its LSTM, or receive information from agent 2 in a communication channel, in which case we could expect the following approximation to be valid
$$
Q^\pi(\bs,\ba) =: \bar{Q}^\pi_1(\bs,\ba) + \bar{Q}^\pi_2(\bs,\ba)  \approx \tilde Q^\pi_1(h^1,a^1) + \tilde Q^\pi_2(h^2,a^2) 
$$
Our architecture therefore encourages this decomposition into simpler functions, if possible. We see that natural decompositions of this type arise in practice (see Section~\ref{LearnedQSection}).

One approach to reducing the number of learnable parameters, is to share certain network weights between agents.
Weight sharing also gives rise to the concept of agent invariance, which is useful for avoiding the lazy agent problem.

\begin{definition}[Agent Invariance]
If for any permutation (bijection) $p:\{1,...,d\}\to\{1,...,d\}$, 
$$\pi(p(\bar{h}))=p(\pi(\bar{h}))$$
we say that $\pi$ is \emph{agent invariant}. 
\end{definition}

It is not always desirable to have agent invariance, when for example specialized roles are required to optimize a particular system. In such cases we provide each agent with \emph{role information}, or an identifier. The role information is provided to the agent as a 1-hot encoding of their identity concatenated with every observation at the first layer. When agents share all network weights they are then only \emph{conditionally agent invariant}, i.e.\ have identical policies only when conditioned on the same role. We also consider information channels between agent networks, i.e. differentiable connections between agent network modules. These architectures, with shared weights, satisfy agent invariance.

\section{Experiments} \label{ExperimentSection}
We introduce a range of two-player domains, and experimentally evaluate the introduced value-decomposition agents with different levels of enhancements, evaluating each addition in a logical sequence.  We use two centralized agents as baselines, one of which is introduced here again relying on learned value-decomposition, as well as an individual agent learning directly from the joint reward signal. We perform this set of experiments on the same form of two dimensional maze environments used by \citet{leibo2017}, but with different tasks featuring more challenging coordination needs. Agents have a small $3\times5\times5$ observation window, the first dimension being an RGB channel, the second and third are the maze dimensions, and each agent sees a box 2 squares either side and 4 squares forwards, see Figures~\ref{Independent} and \ref{Value-Decomp}. The simple graphics of our domains helps with running speed while, especially due to their multi-agent nature and severe partial observability and aliasing (very small observation window combined with map symmetries), they still pose a serious challenge and is comparable to the state-of-the-art in multi-agent reinforcement learning \citep{leibo2017}, which exceeds what is common in this area \citep{TuylsW12}.

\subsection{Agents}
Our agent's learning algorithm is based on DQN \citep{dqn15} and includes its signature techniques of experience replay and target networks, enhanced with an LSTM value-network as in \citet{HausknechtStoneRecurrent} (to alleviate severe partial observability), learning with truncated back-propagation through time, multi-step updates with forward view eligibility traces \citep{HarbPrecupRecurrance} (which helps propagating learning back through longer sequences) and the dueling architecture \citep{WangDuelling} (which speeds up learning by generalizing across the action space). Since observations are from a local perspective, we do not benefit from convolutional networks, but use a fully connected linear layer to process the observations. 

Our network architectures first process the input using a fully connected linear layer with $32$ hidden units followed by a ReLU layer, and then an LSTM, with $32$ hidden units followed by a ReLU layer, and finally a linear dueling layer, with $32$ units. This produces a value function $V(s)$ and \emph{advantage function} $A(s,a)$, which are combined to compute a $Q$-function $Q(s,a) = V(s) + A(s,a)$ as described in \citet{WangDuelling}. Layers of $32$ units are sufficiently expressive for these tasks with limited observation windows.

The architectures (see Appendix B for detailed diagrams) differ between approaches by what is input into each layer. For architectures without centralization or information channels, one observation of size $3\times 5\times 5$ is fed to the first linear layer of $32$ units, followed by the ReLU layer and the LSTM (see Figure~\ref{Independent}). For the other (information channels and centralized) agents, $d$ such observations are fed separately to identical such linear layers and then concatenated into $64$ dimensional vectors before passing though ReLUs to an LSTM. 

For architectures with information channels we concatenate the outputs of certain layers with those of other agents. To preserve agent invariance, the agent's own previous output is always included first. For low-level communication, the signal's concatenation is after the first fully connected layer, while for high-level communication the concatenation takes place on the output of the LSTM layer. Note, that this has the implication that what starts as one agent's gradients are back-propagated through much of the other agents network, optimizing them to serve the purposes of all agents. Hence, representing in that sense, a higher degree of centralization than the lower-level sharing.


We have found a trajectory length of $8$, determining both the length of the forward view and the length of the back propagation through time is sufficient for these domains. We use an eligibility trace parameter $\lambda=0.9$. In particular, the individual agents learning directly from the joint reward without decomposition or information channels, has worse performance with lower $\lambda$. The Adam \citep{Kingma14} learning rate scheme initialized with $0.0001$ is uniformly used, and further fine-tuning this per agent (not domain) does not dramatically change the total performance. The agents that we evaluate are listed in the table above.

\begin{SCtable}
 \begin{tabular}{c c c c c c c}
 \toprule
 Agent & V. & S. & Id & L. & H. & C. \\ [0.5ex] 
\midrule
 1 &   &  &   &  &   &  \\
 2 & \checkmark &   &   &   &   &  \\
 3 & \checkmark & \checkmark &   &   &   &  \\
 4 & \checkmark & \checkmark & \checkmark &   &   &  \\
 5 & \checkmark & \checkmark & \checkmark & \checkmark &   &  \\
 6 & \checkmark & \checkmark & \checkmark &   & \checkmark &  \\
 7 & \checkmark & \checkmark & \checkmark & \checkmark & \checkmark &  \\
 8 & \checkmark &   &   &   &   & \checkmark\\
 9 &   &   &   &   &   & \checkmark\\
\bottomrule
\end{tabular}
\caption{Agent architectures. V is value decomposition, S means shared weights and an invariant network, Id means role info was provided, L stands for lower-level communication, H for higher-level communication and C for centralization. These architectures were selected to show the advantages of the independent agent with value-decomposition and to study the benefits of additional enhancements added in a logical sequence.}
\label{tab:archs}
\end{SCtable}

\subsection{Environments}

\begin{figure}
\begin{minipage}{.5\textwidth}
\centering
\includegraphics[height=1.7cm]{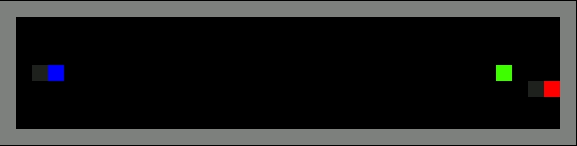}
\vspace{0.05cm}
\includegraphics[height=1.7cm, width=6.75cm]{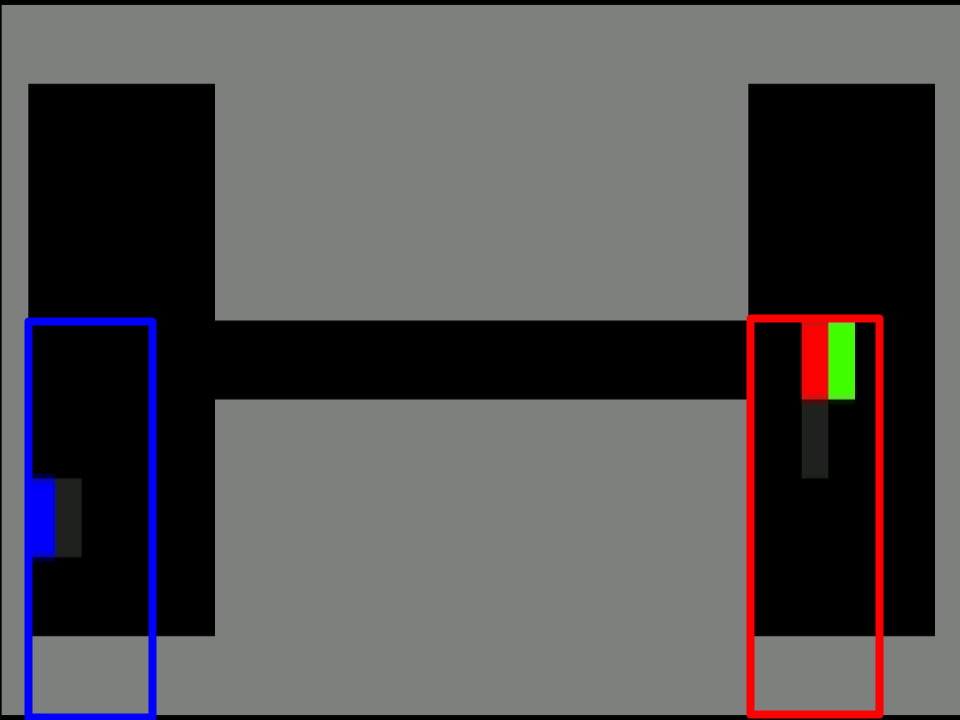}

\end{minipage}
\begin{minipage}{.5\textwidth}
\centering
\includegraphics[height=1.7cm]{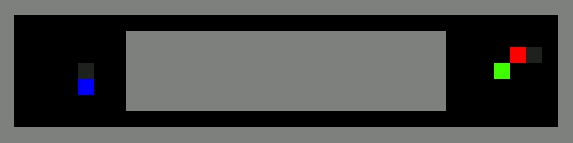}
\vspace{0.05cm}
\includegraphics[height=1.7cm, width=6.75cm]{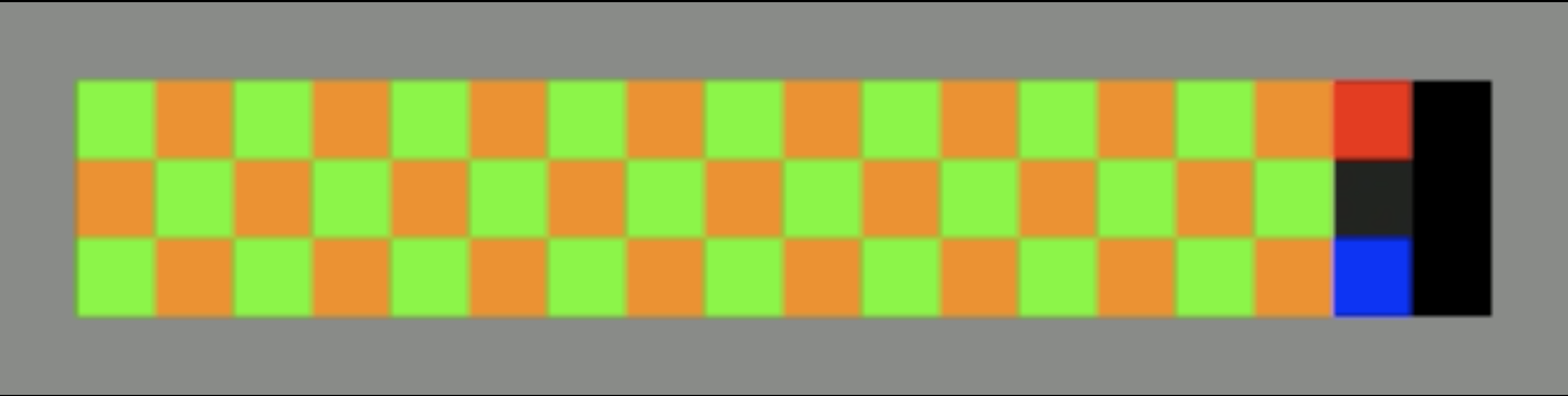} 
\end{minipage}
\caption{Maps for Fetch and Switch: open map (top left), map with 1 corridor (bottom left) and 2 corridors (top right). The green square is the goal for the agent to the left (blue). A similar goal is seen for the other agent (red) to the left but not displayed. The agents' observation windows are shown in the bottom left. Bottom right is the map for Checkers. Lemons are orange, apples green and agents red/blue.} \label{S0}
\end{figure}

We use 2D grid worlds with the same basic functioning as \citet{leibo2017}, but with different tasks we call Switch, Fetch and Checkers. We have observations of byte values of size $3\times5\times5$ (RGB), which represent a window depending on the player’s position and orientation by extending $4$ squares ahead
and $2$ squares on each side. Hence, agents are very short-sighted. The actions are: step forward, step backward,
step left, step right, rotate left, rotate right, use beam
and stand still. The beam has no effect in our games, except for lighting up a row or column of squares straight ahead with yellow. 
Each player appears as a  blue square in its own observation, and the other player, when in the observation window, is shown in red for Switch and Escape, and lighter blue for Fetch. We use three different maps shown in Figure~\ref{S0} for both Fetch and Switch and a different one for Checkers, also shown in Figure~\ref{S0} (bottom right). The tasks repeat as the agents succeed (either by full reset of the environment in Switch and Checkers or just by pickup being available again in Fetch), in training for 5,000 steps and 2,000 in testing.

{\bf Switch:} The task tests if two agents can effectively coordinate their use of available routes on two maps with narrow corridors. The task is challenging because of strong observation aliasing. The two agents appear on different ends of a map, and must reach a goal at the other end. If agents collide in a corridor, then one agent needs to leave the corridor to allow the other to pass. When both players have reached their goal the environment is reset. A point is scored whenever a player reaches a goal. 

{\bf Fetch:} The task tests if two agents can synchronize their behaviour, when picking up objects and returning them to a drop point. In the Fetch task both players start on the same side of the map and have pickup points on the opposite side. A player scores $3$ points for the team for pick-up, and another $5$ points for dropping off the item at the drop point near the starting position. Then the pickup is available to either player again. It is optimal for the agents to cycle such that when one player reaches the pickup point the other returns to base, to be ready to pick up again. 



{\bf Checkers:} 
The map contains apples and lemons. The first player is very sensitive and scores $10$ for the team for an apple (green square) and $-10$ for a lemon (orange square). The second, less sensitive player scores $1$ for the team for an apple and $-1$ for a lemon. There is a wall of lemons between the players and the apples. Apples and lemons disappear when collected, and the environment resets when all apples are eaten. It is important that the sensitive agent eats the apples while the less sensitive agent should leave them to its team mate but clear the way by eating obstructing lemons.

\subsection{Results}

\begin{figure*}[t]
\includegraphics[width=14.cm]{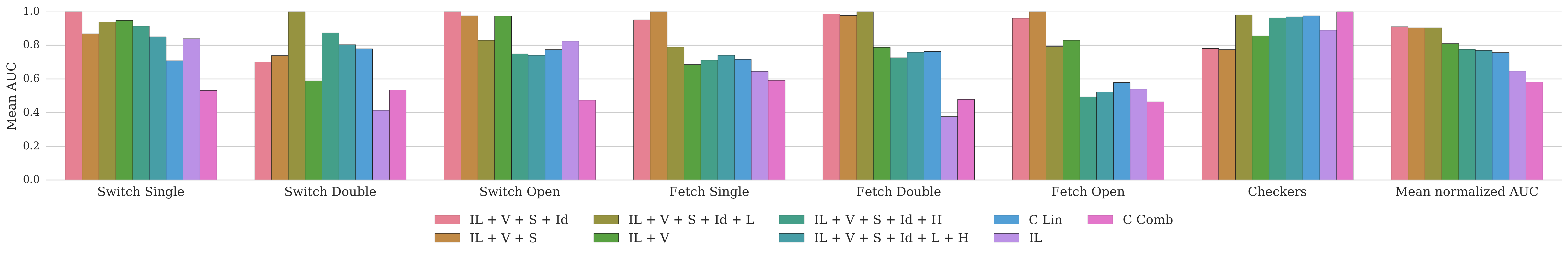}
\caption{Barplots showing normalized AUC for each agent and domain over 50000 episodes of training and the mean across domains.}\label{barplot_all}
\end{figure*}

\begin{figure}
\begin{minipage}{.5\textwidth}
\includegraphics[width =8.0cm]{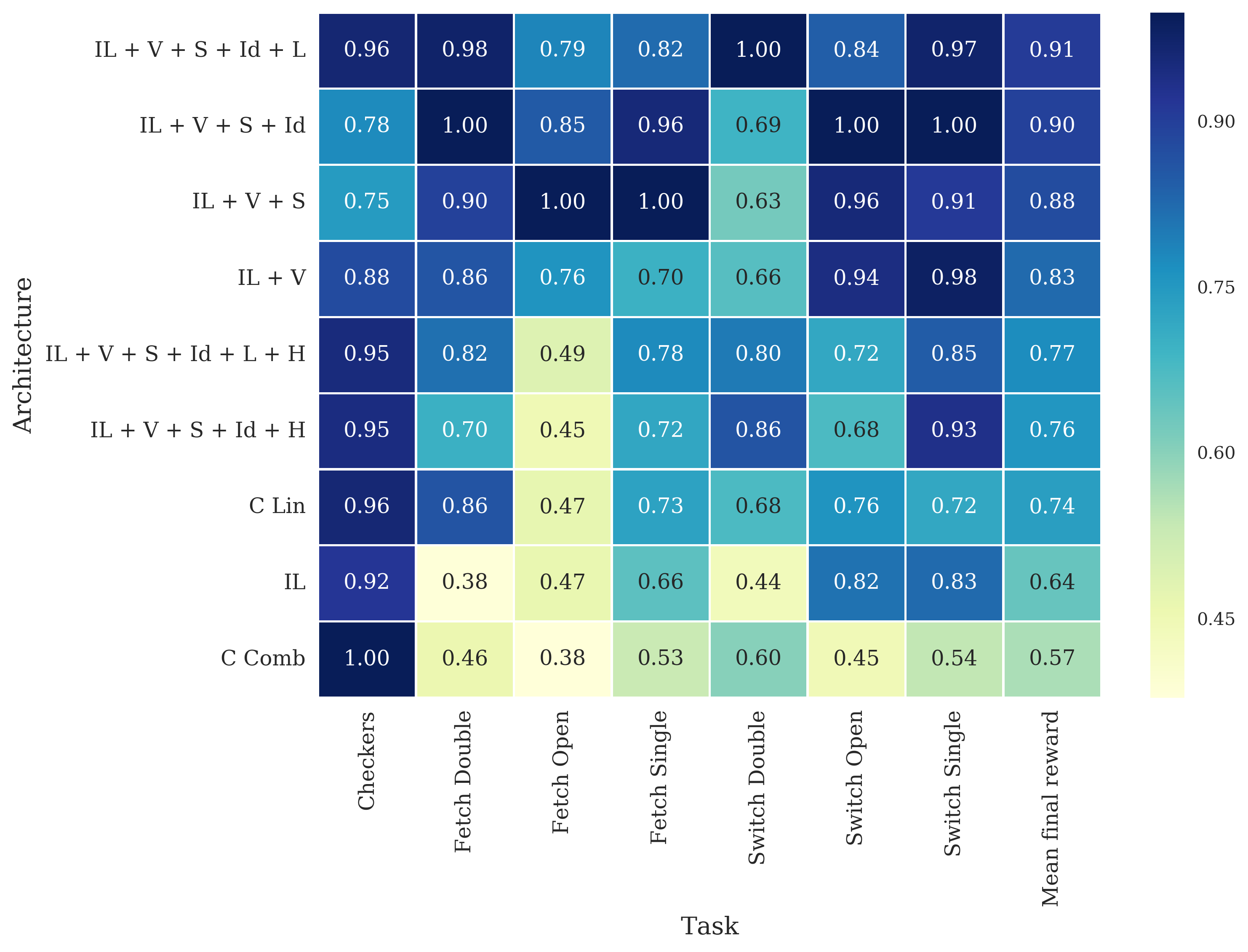}
\caption{Heatmap showing each agent's final performance, averaged over the last 5,000 episodes of 50,000 and across ten runs, normalized by the best architecture per task. The agents are ordered according to average over the domains, which can be seen in the right most column. Value-Decomposition architecture strongly outperform Individual Learners and Centralization} \label{heat}

\end{minipage}
\begin{minipage}{.5\textwidth}
\includegraphics[width=7.0cm]{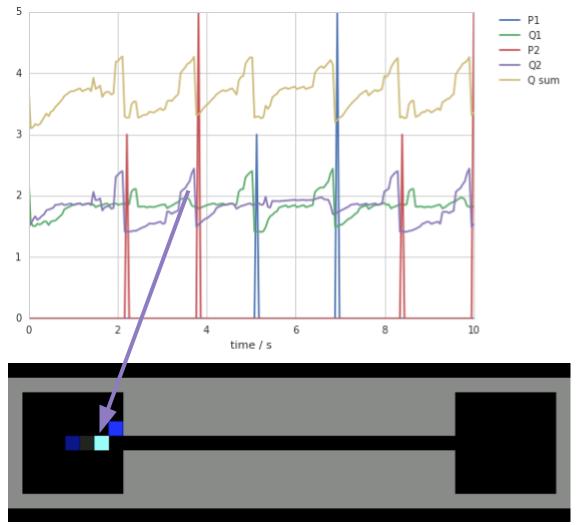}
\caption{The learned $Q$-decomposition in Fetch. The plot shows the total $Q$-function (yellow), the value of agent~1 (green), the value of agent~2 (purple), rewards from agent~1 (blue) events and agent~2 (red). Highlighted is a situation in which agent~2's $Q$-function spikes (purple line), anticipating reward for an imminent drop-off. The other agent's $Q$-function (green) remains relatively flat.}\label{Qplot}
\end{minipage}

\end{figure}

We compare the eight approaches listed in Table~\ref{tab:archs}, on the seven tasks. Each is run ten times, with different random seeds determining spawn points in the environment, as well as initializations of the neural networks. 
We calculated curves of the average performance over 50,000 episodes (plots in Appendix A) for each approach on each task and we display the normalized area under the curve in Figure \ref{barplot_all}. Figure \ref{heat} displays the normalized final performance averaged over runs and the last 1,000 episodes. Average performance across tasks is also shown for both ways of evaluation.

The very clear conclusion is that architectures based on value-decomposition perform much better, with any combination of other techniques or none, than the centralized approach and individual learners. The centralized agent with value-decomposition is better than the combinatorially centralized as well as individual learners while worse than the more individual agents with value-decomposition.

We particularly see the benefit of shared weights on the hard task of Fetch with one corridor. Without sharing, the individual value-decomposition agent suffers from the lazy agent problem.  The agent with weight sharing and role information also perfectly learns the one corridor Fetch task. It performs better than the agent just sharing weights on Switch, where coordination, in particular with one corridor, is easier with non-identical agents. Further, shared weights are problematic for the Checkers task because the magnitude of rewards (and hence the value function) from one agent is ten times higher than for the other agent. 

Adding information channels does increase learning complexity because the input comes from more than one agent. However, the checkers task, designed for the purpose, shows that it can be very useful. Overall, the low-level channels where the agent's LSTM processes the combined observations of both agents turned out to learn faster in our experiments than the more centralized high level communication (after the LSTM). 

\subsection{The Learned $Q$-Decomposition}\label{LearnedQSection}

Figure~\ref{Qplot} shows the learned $Q$-decomposition for the value-decomposition network, using shared weights, in the game of Fetch. A video of the corresponding game can be seen at \citet{B17}. Spikes correspond to pick-up events (short spikes, 3 reward points), and return events (large spikes, 5 reward points). These are separated into events due to agent~1 (blue spikes) and agent~2 (red spikes). This disambiguation is for illustration purposes only: the environment gives a reward to the whole team for all of these events. The total $Q$-function is seen in yellow, clearly anticipating the team reward events, and dropping shortly afterwards. The component $Q$-functions $\tilde Q_1$ and $\tilde Q_2$ for agents~1 and 2 are shown in green and purple. These have generally disambiguated the $Q$-function into rewarding events separately attributable to either player. The system has learned to autonomously decompose the joint $Q$-function into sensible components which, when combined, result in an effective $Q$-function. This would be difficult for independent learners since many rewards would not be observed by both players, see e.g. the situation at 15-16 seconds in the corresponding video available at \citet{B17}.








\section{Conclusions}
We study cooperative multi-agent reinforcement learning where only a single joint reward is provided to the agents. We found that the two naive approaches, individual agents learning directly from team reward, and fully centralized agents, provide unsatisfactory solutions as previous literature has found in simpler environments, while our value-decomposition networks do not suffer from the same problems and shows much better performance across a range of more complex tasks. Further, the approach can be nicely combined with weight sharing and information channels, leading to agents that consistently optimally solve our new benchmark challenges. 



Value-decomposition networks are a step towards automatically decomposing complex learning problems into local, more readily learnable sub-problems. In future work we will investigate the scaling of value-decomposition with growing team sizes, which make individual learners with team reward even more confused (they mostly see rewards from other agents actions), and centralized learners even more impractical. We will also investigate decompositions based on non-linear value aggregation.

\FloatBarrier


\newpage

\section*{Appendix A: Plots}

\begin{figure}[h!]
\centering
\includegraphics[width=8.0cm]{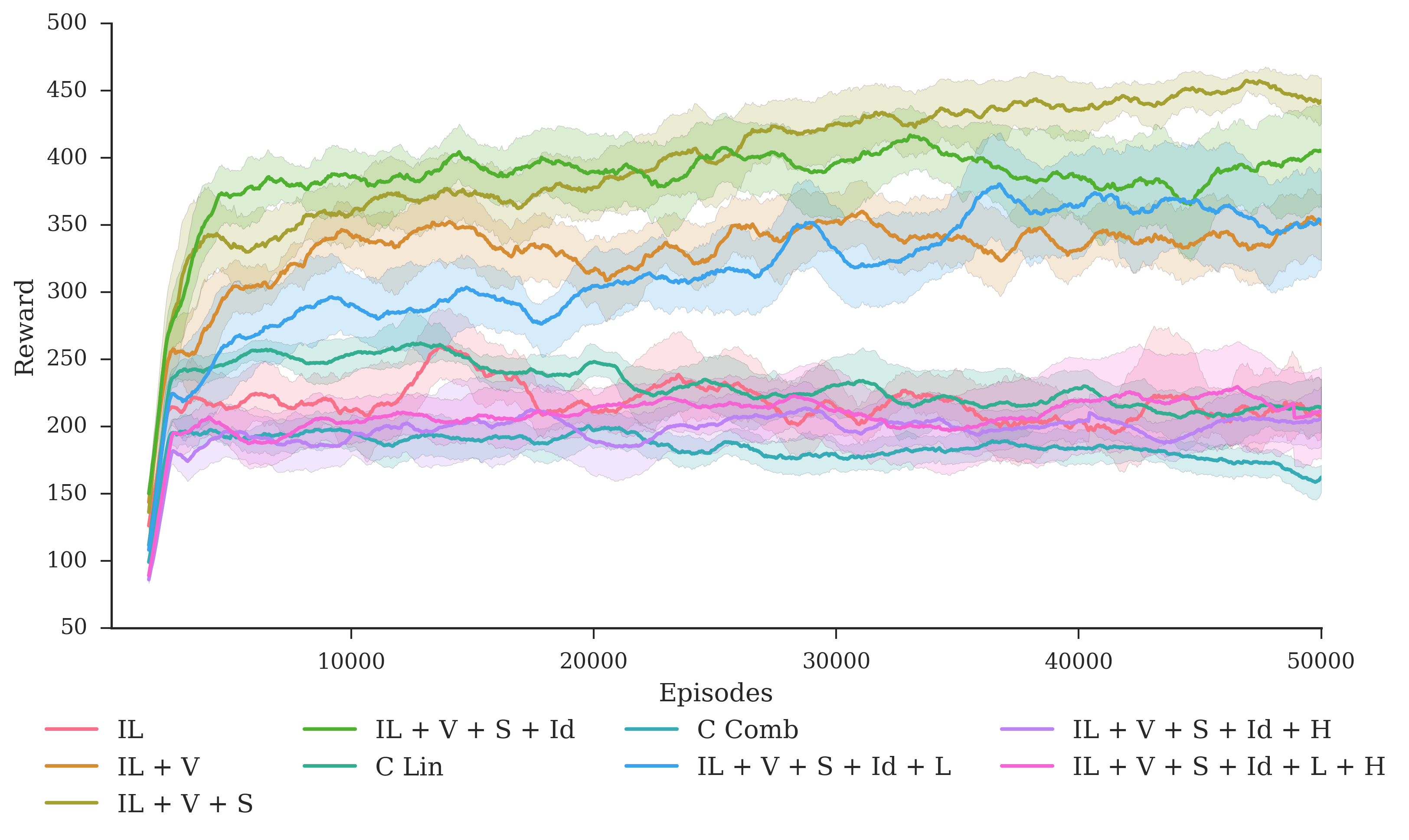}
\caption{Average reward with 90\% confidence intervals for ten runs of the nine architectures on the Fetch domain with the open map}
\end{figure}
\begin{figure}[h!]
\centering
\includegraphics[width=8.0cm]{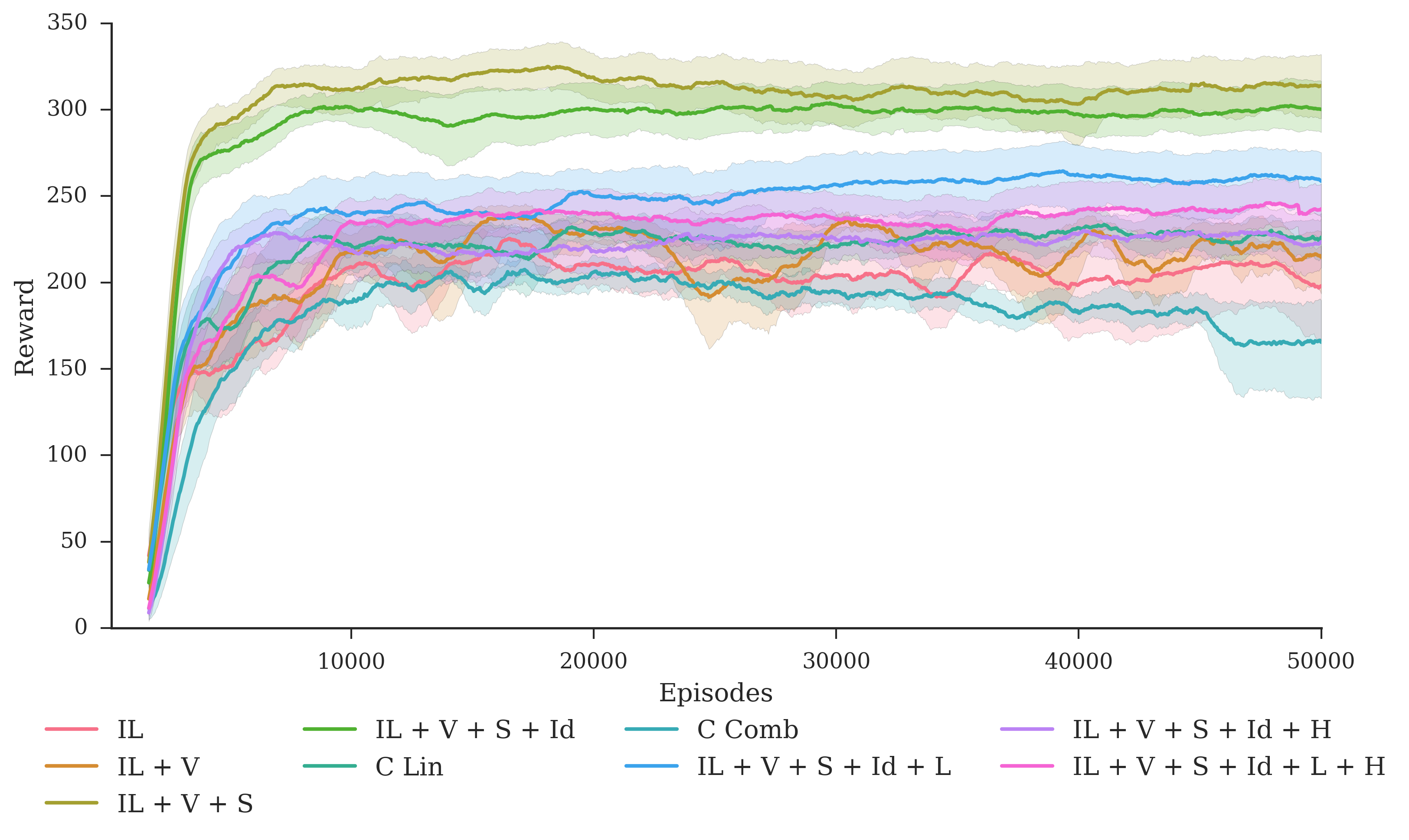}
\caption{Average reward with 90\% confidence intervals for ten runs of the nine architectures on the Fetch domain with one corridor}
\end{figure}
\begin{figure}[h!]
\centering
\includegraphics[width=8.0cm]{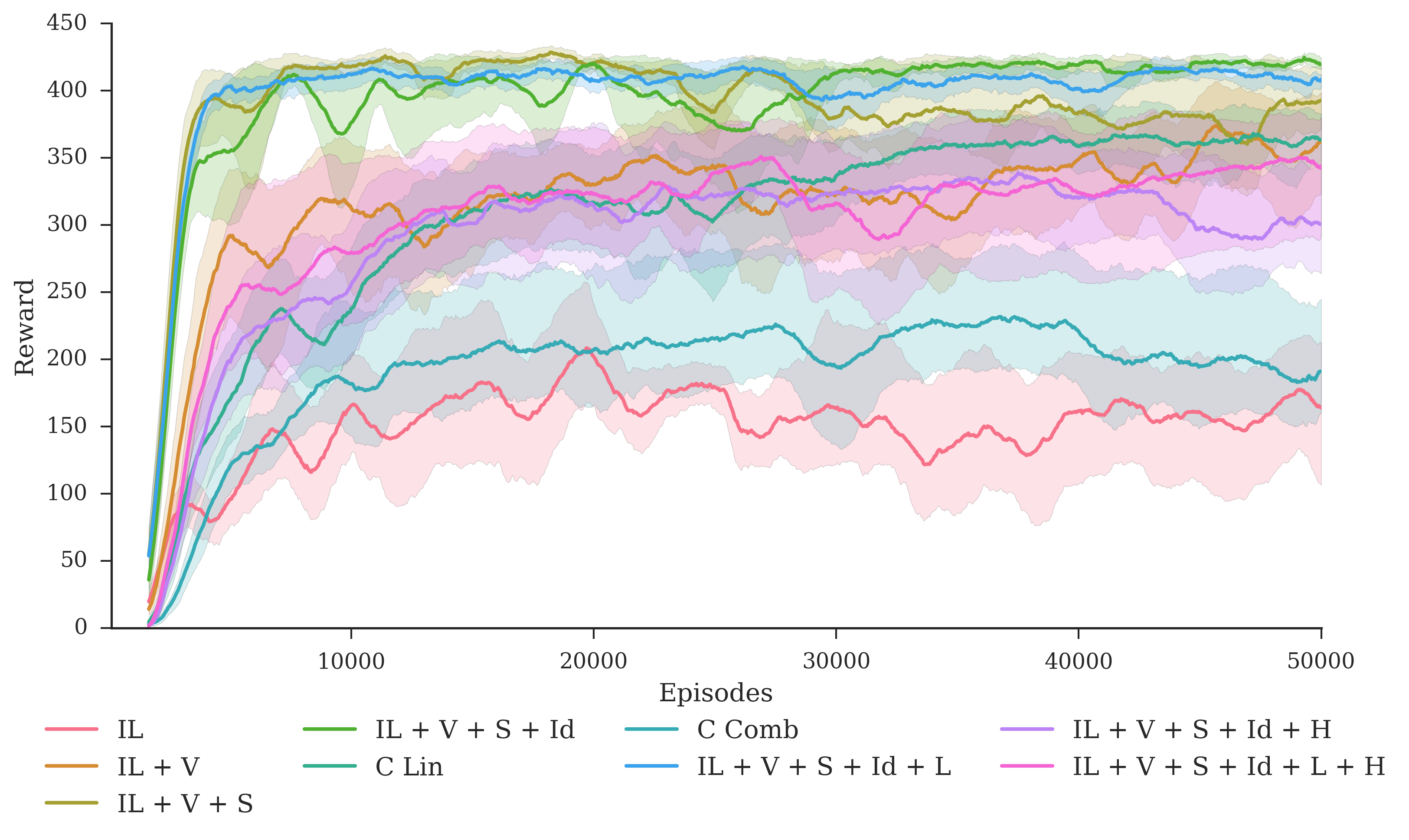}
\caption{Average reward with 90\% confidence intervals for ten runs of the nine architectures on the Fetch domain with two corridors}
\end{figure}
\begin{figure}[h!]
\centering
\includegraphics[width=8.0cm]{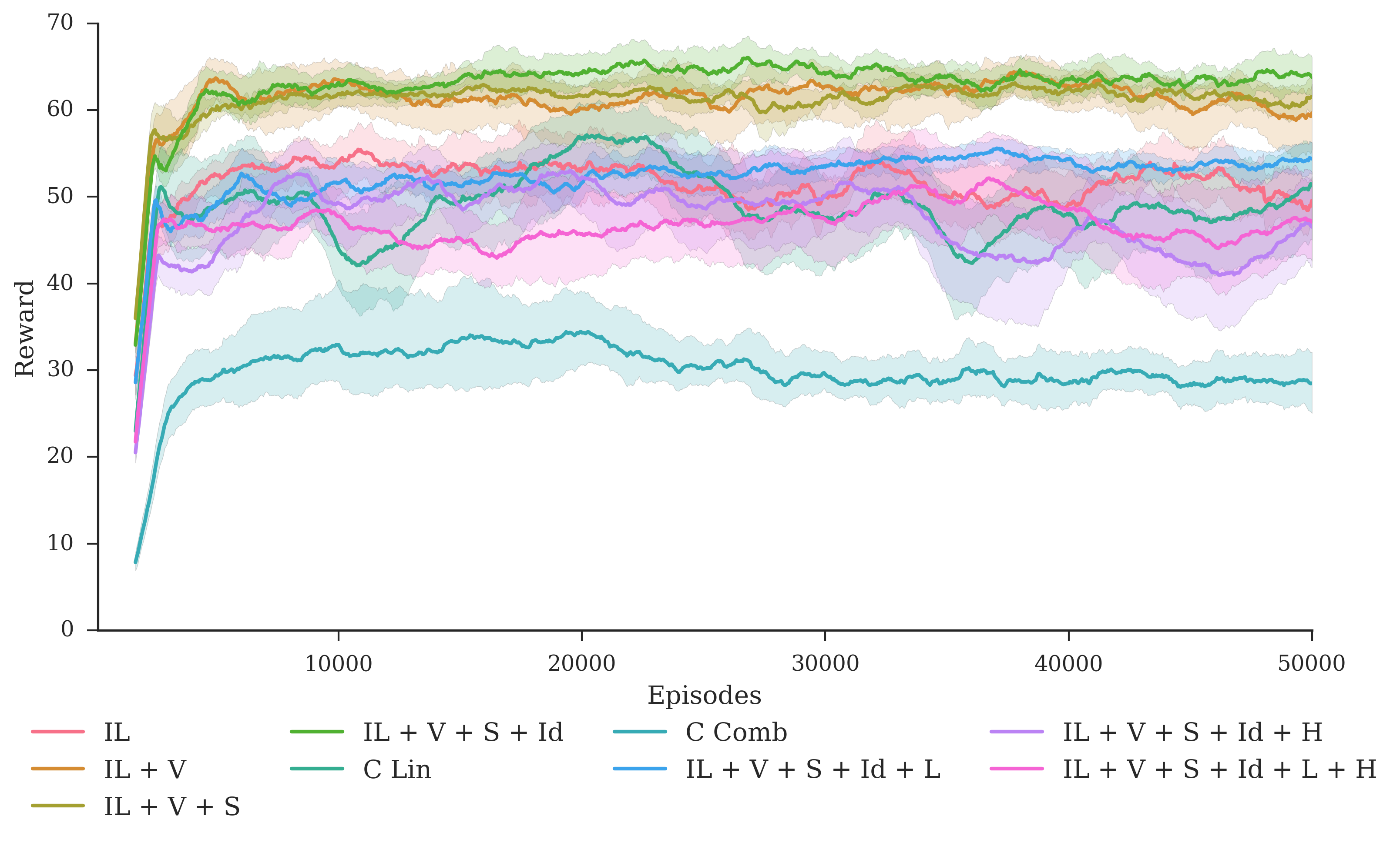}
\caption{Average reward with 90\% confidence intervals for ten runs of the nine architectures on the Switch domain with the open map}
\end{figure}
\begin{figure}[h!]
\centering
\includegraphics[width=8.0cm]{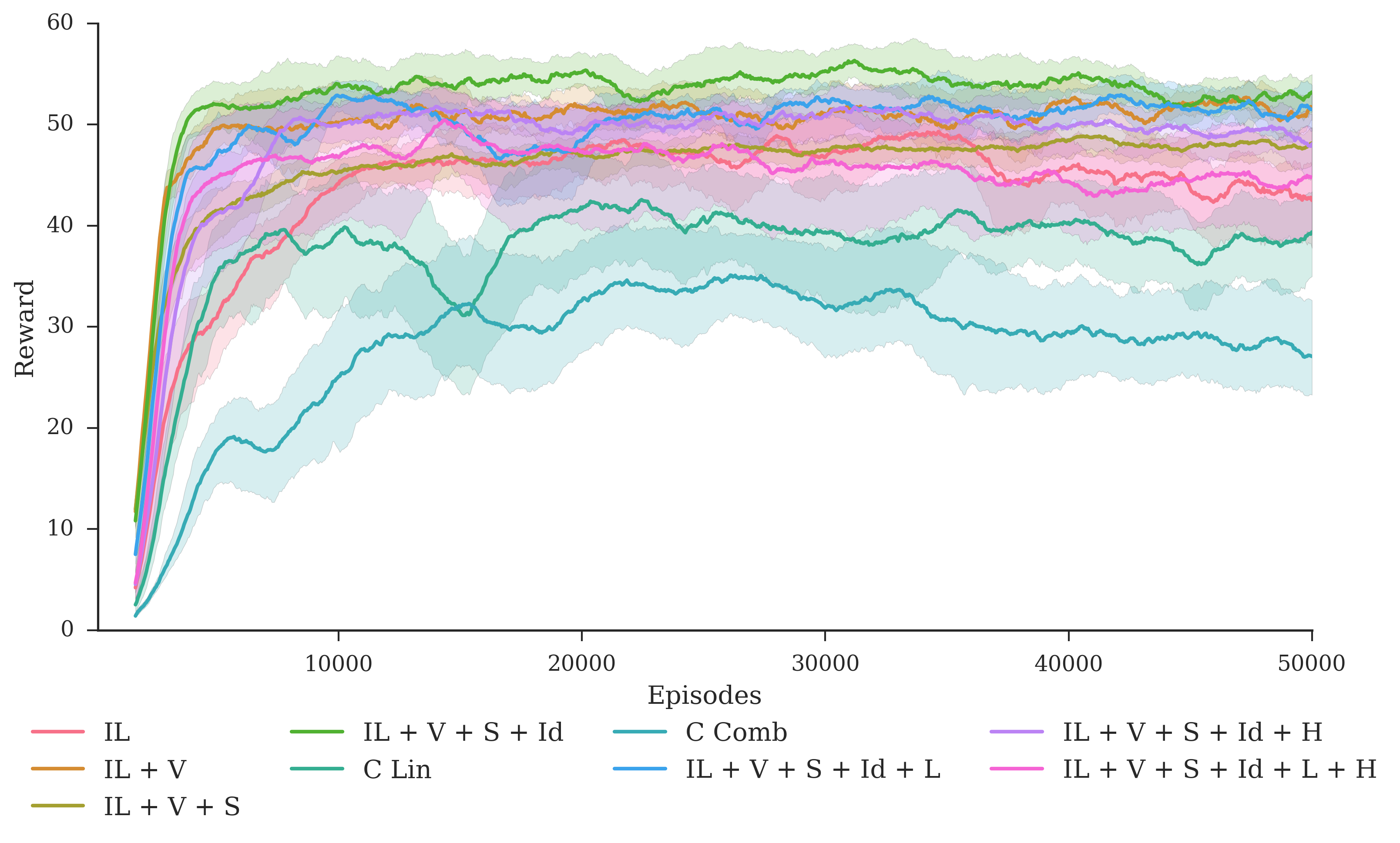}
\caption{Average reward with 90\% confidence intervals for ten runs of the nine architectures on the Switch domain with one corridor}
\end{figure}

\begin{figure}[h!]
\centering
\includegraphics[width=8.0cm]{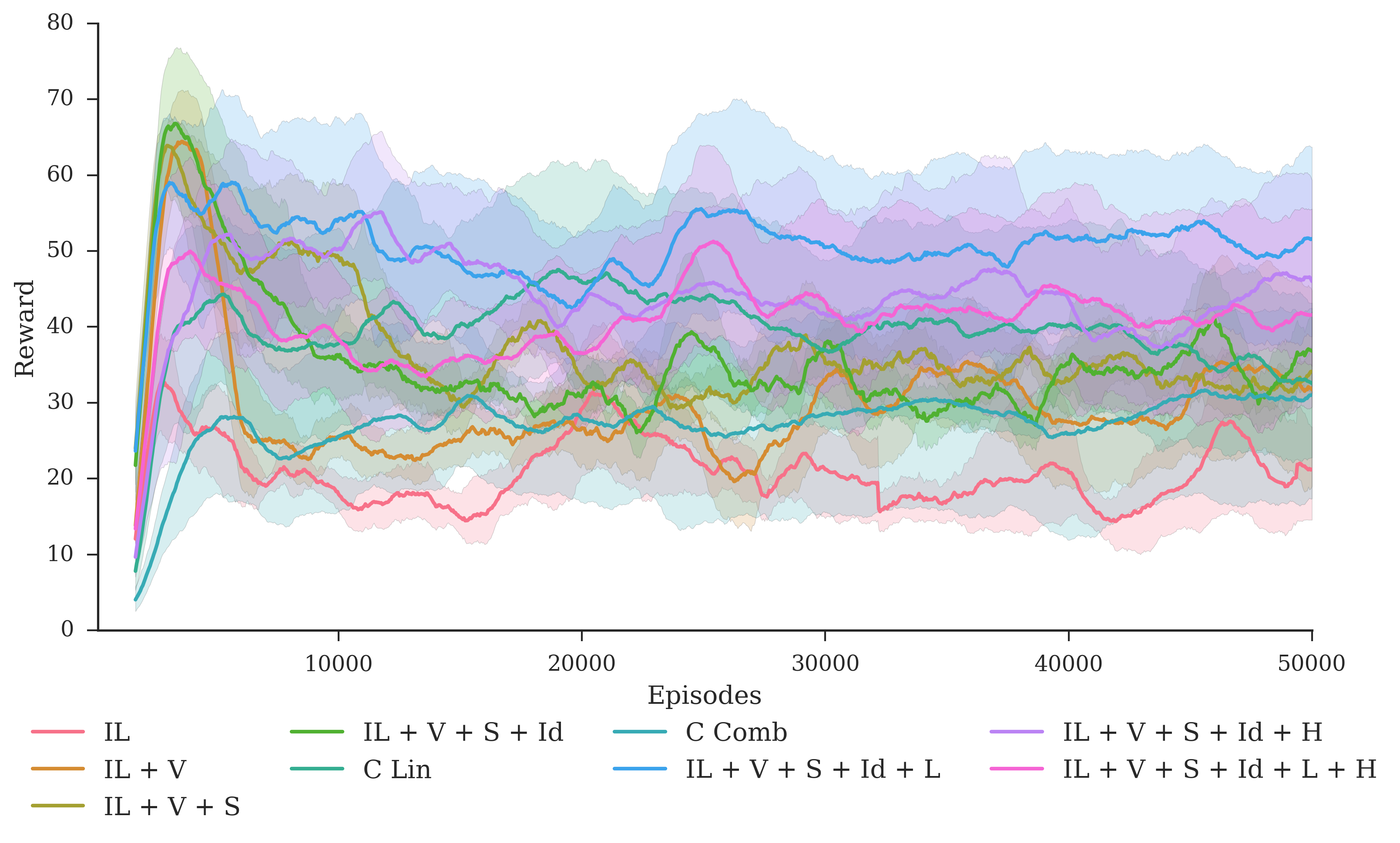}
\caption{Average reward with 90\% confidence intervals for ten runs of the nine architectures on the Switch domain with two corridors}
\end{figure}

\begin{figure}[h!]
\centering
\includegraphics[width=8.0cm]{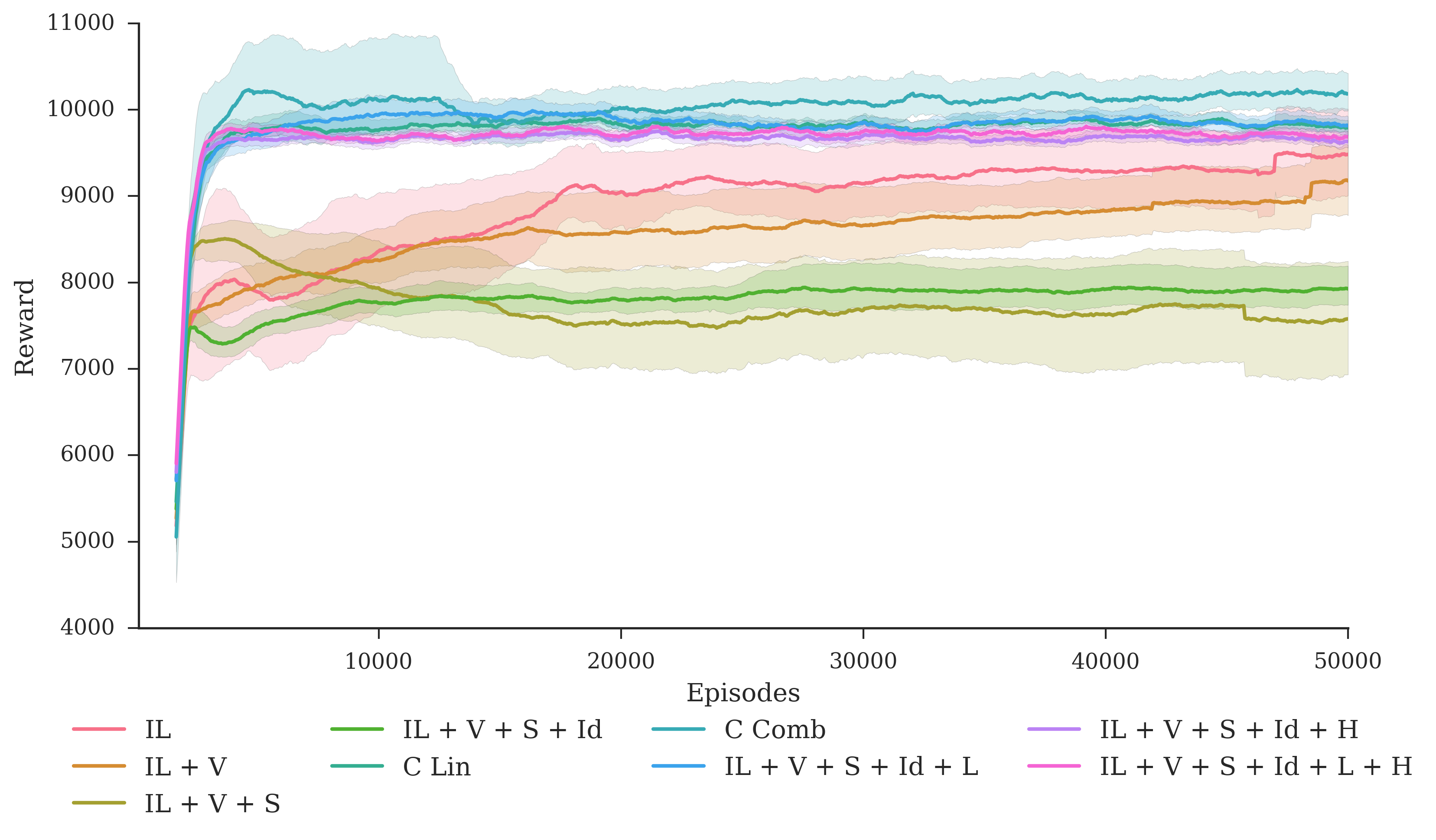}
\caption{Average reward with 90\% confidence intervals for ten runs of the nine architectures on the Checkers domain}
\end{figure}




\FloatBarrier
\section*{Appendix B: Diagrams}

\begin{figure}[h!]
 \centering
\includegraphics[height=6.0cm]{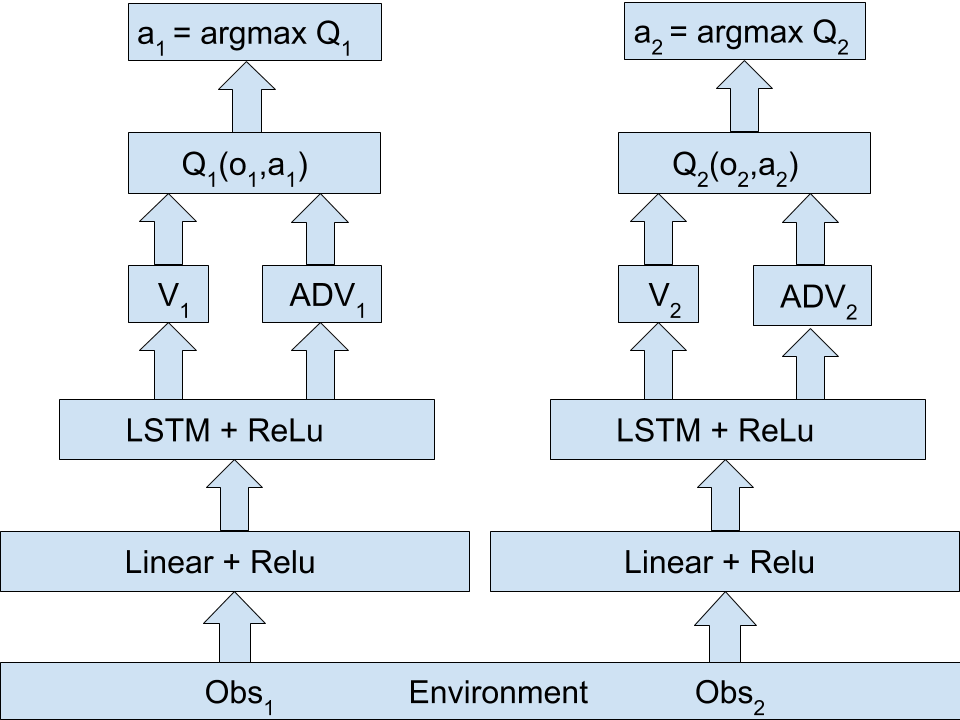}
\caption{Independent Agents Architecture}

\end{figure}

\begin{figure}[h!]
 \centering
\includegraphics[height=6.0cm]{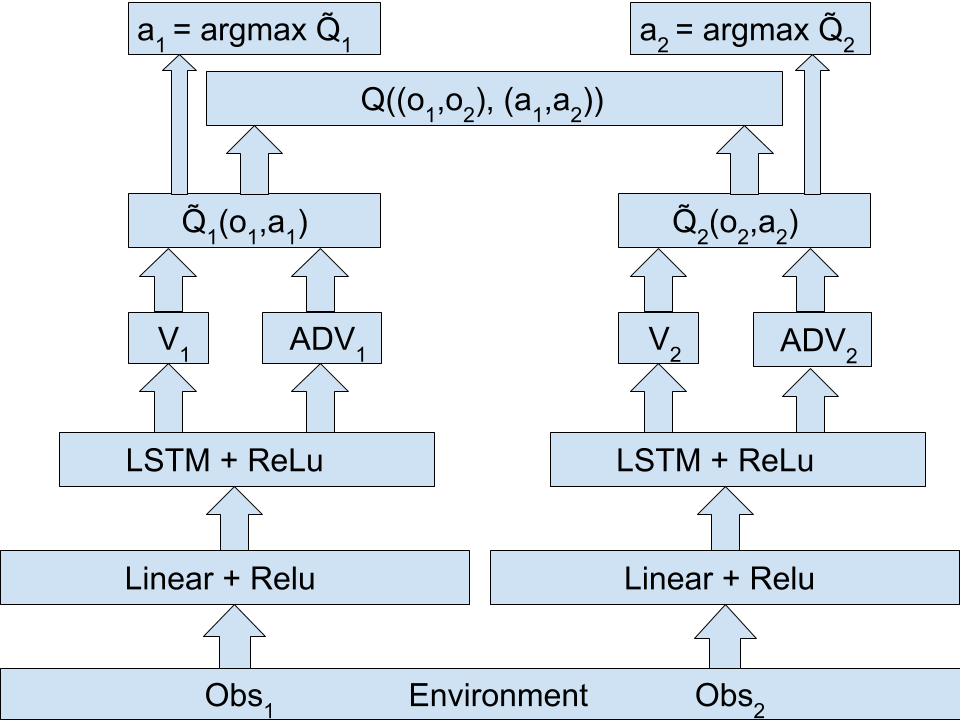}
\caption{Value-Decomposition Individual Architecture}

\end{figure}

\begin{figure}[h!]
 \centering
\includegraphics[height=6.0cm]{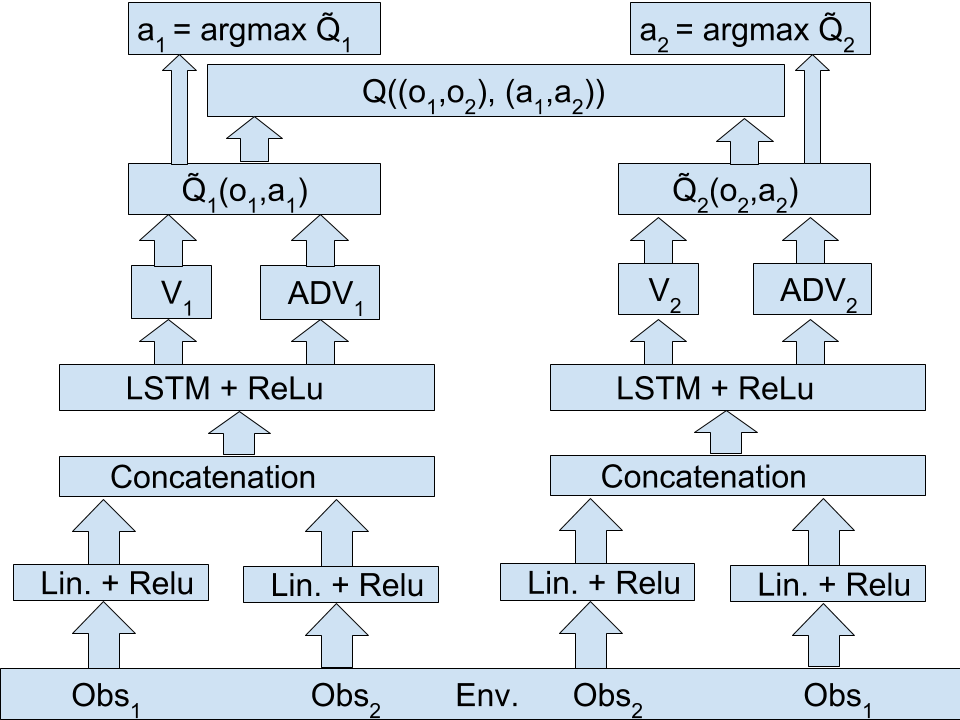}
\caption{Low-level communication Architecture}

\end{figure}

\begin{figure}[h!]
 \centering
\includegraphics[height=6.0cm]{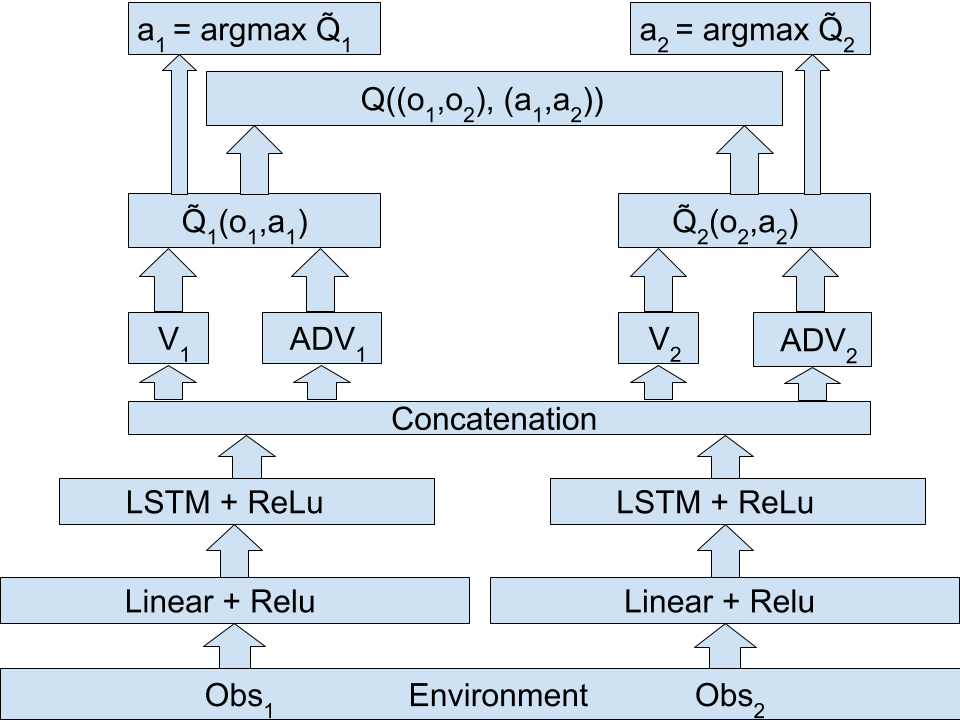}
\caption{High-level communication Architecture}

\end{figure}

\begin{figure}[h!]
 \centering
\includegraphics[height=6.0cm]{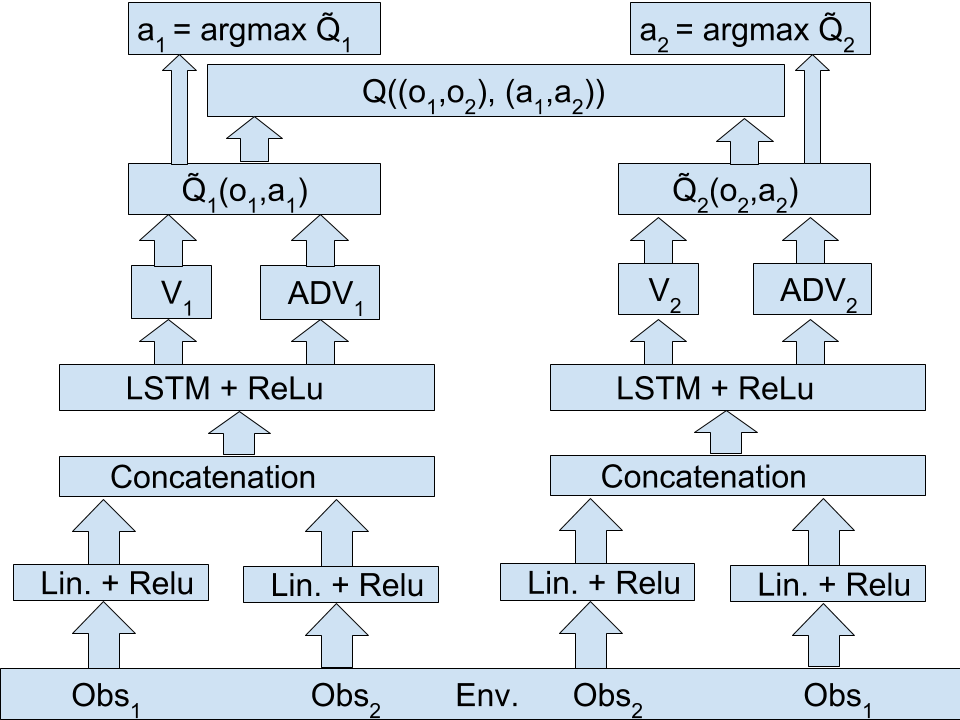}
\caption{Low-level communication Architecture}

\end{figure}

\begin{figure}[h!]
 \centering
\includegraphics[height=6.0cm]{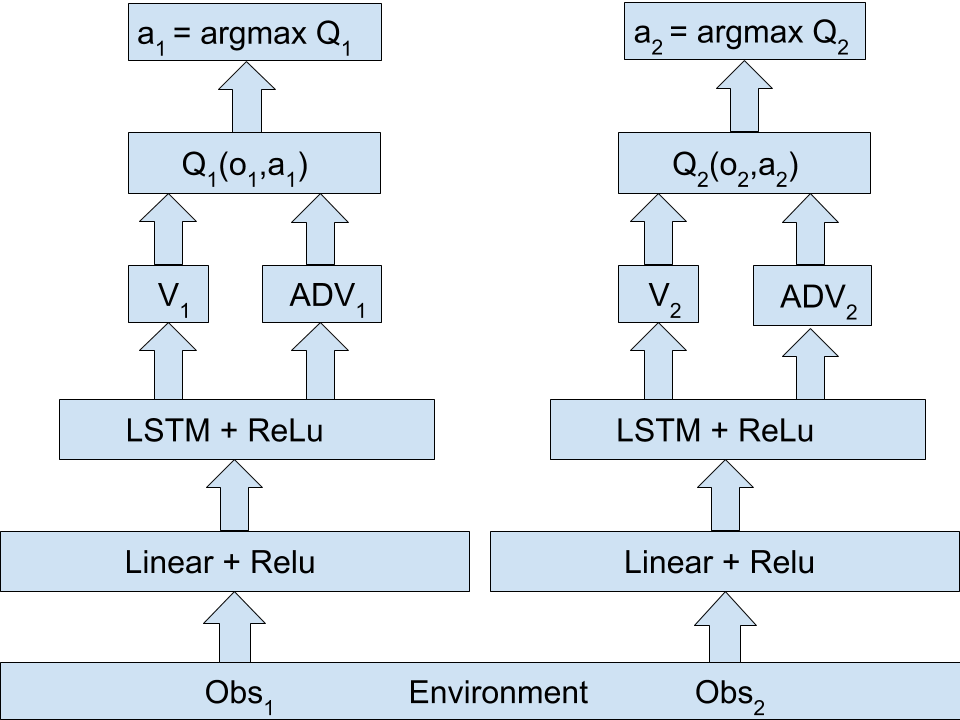}
\caption{Independent Agents Architecture}

\end{figure}

\begin{figure}[h!]
 \centering
\includegraphics[height=6.0cm]{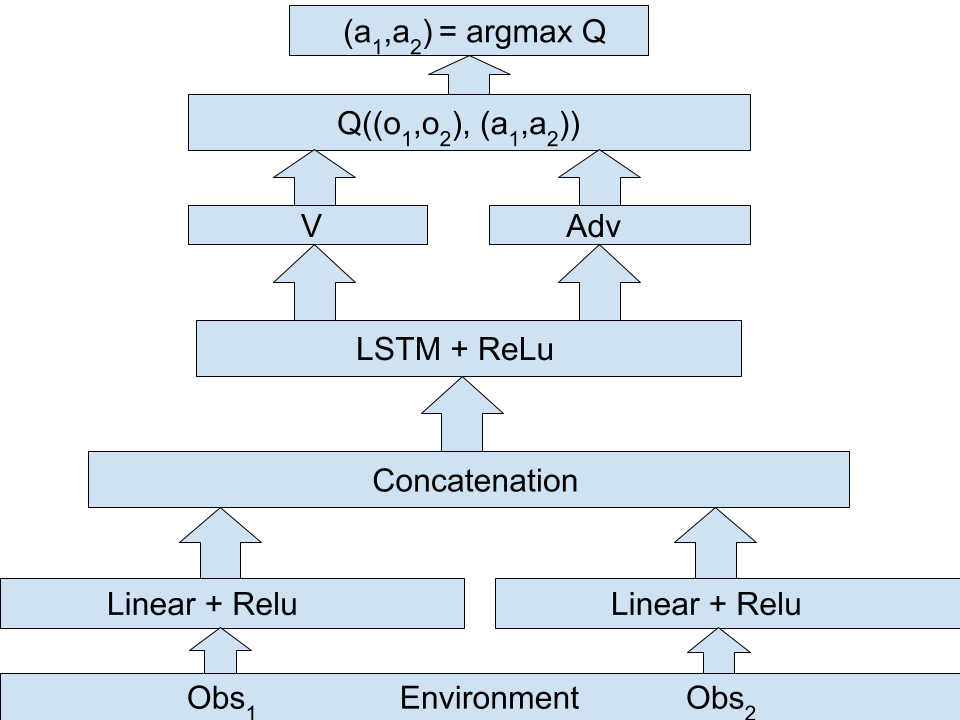}
\caption{Combinatorially Centralized Architecture}

\end{figure}

\begin{figure}[h!]
 \centering
\includegraphics[height=6.0cm]{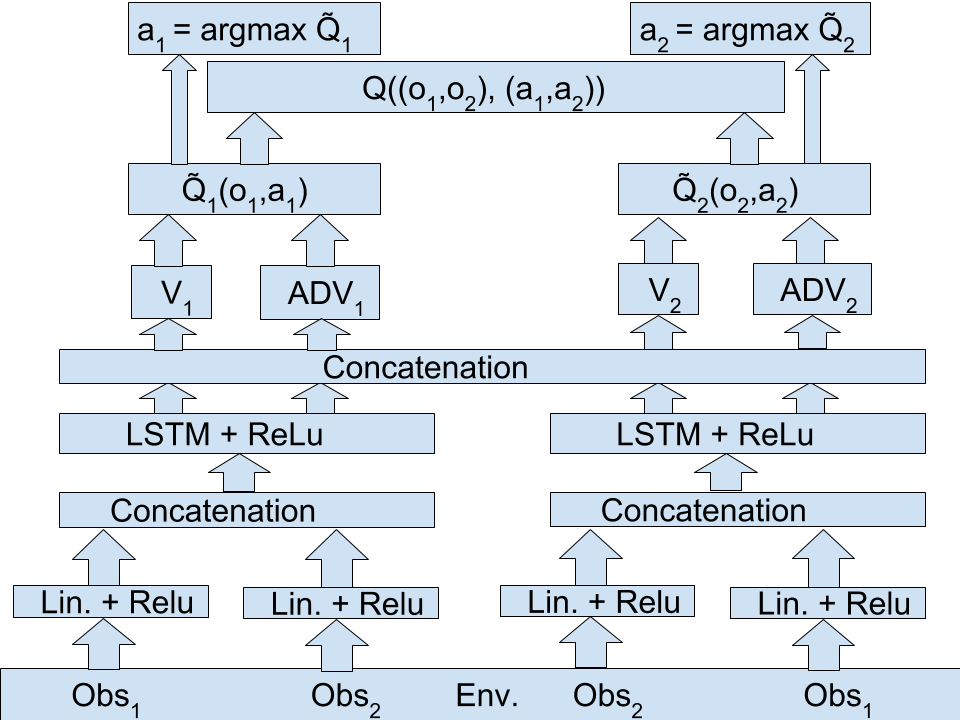}
\caption{High+Low-level communication Architecture}

\end{figure}

\end{document}